\documentclass{article}

    \PassOptionsToPackage{numbers, compress}{natbib}

\usepackage[preprint]{neurips_2023}

\usepackage[utf8]{inputenc} 
\usepackage[T1]{fontenc}    
\usepackage{hyperref}       
\usepackage{url}            
\usepackage{booktabs}       
\usepackage{amsfonts}       
\usepackage{nicefrac}       
\usepackage{microtype}      
\usepackage{xcolor}         
\usepackage{tabularx, makecell, multirow} 
\usepackage{adjustbox}
\usepackage{graphicx}
\usepackage{subfigure}
\usepackage{multicol}
\usepackage{amsmath}
\usepackage{amsthm}
\usepackage{bm}

\definecolor{darkgreen}{rgb}{0,0.5,0}
\hypersetup{colorlinks=true, citecolor=darkgreen}
\usepackage{enumitem} 
\usepackage{adjustbox} 
\usepackage{multirow} 
\usepackage{bbm} 
\usepackage{mathtools} 
\usepackage{wrapfig} 

\usepackage{colortbl} 
\usepackage{arydshln} 

\usepackage{pifont} 
\usepackage{graphicx}
\usepackage{caption}
\usepackage{calc}
\usepackage{lipsum}
\usepackage{authblk}

\newcolumntype{L}[1]{>{\raggedright\let\newline\\\arraybackslash\hspace{0pt}}m{#1}}
\newcolumntype{C}[1]{>{\centering\let\newline\\\arraybackslash\hspace{0pt}}m{#1}}

\newcommand\mymodel{DiffusePast}
\newcommand\myloss{Ours-$\mathcal{L}$}

\title{DiffusePast: Diffusion-based Generative Replay for Class Incremental Semantic Segmentation}

\author[1,2]{{Jingfan Chen}}
\author[2,3,4]{{Yuxi Wang}$^{\dagger}$}
\author[1,2]{{Pengfei Wang}}
\author[1,2]{{Xiao Chen}}
\author[2,3,4]{\authorcr {Zhaoxiang Zhang}$^{\dagger}$}
\author[2,3,4]{{Zhen Lei}$^{\dagger}$}
\author[1]{{Qing Li}$^{\dagger}$}

\affil[1]{The Hong Kong Polytechnic University}
\affil[2]{Center for Artificial Intelligence and Robotics, HKISI, CAS}
\affil[3]{Institute of Automation, Chinese Academy of Sciences}
\affil[4]{University of Chinese Academy of Sciences}
\affil[ ]{\texttt{jingfan.chen@connect.polyu.hk}}

\begin{document}

\maketitle

\newcommand{\myfootnote}[1]{%
  \begingroup
  \renewcommand\thefootnote{}\footnotetext{#1}%
  \endgroup
}

\myfootnote{$^{\dagger}$ Equally advising corresponding authors. E-mails: yuxiwang93@gmail.com, zhaoxiang.zhang@ia.ac.cn, zlei@nlpr.ia.ac.cn, csqli@comp.polyu.edu.hk}

\begin{abstract}

The Class Incremental Semantic Segmentation (CISS) extends the traditional segmentation task by incrementally learning newly added classes. Previous work has introduced generative replay, which involves replaying old class samples generated from a pre-trained GAN, to address the issues of catastrophic forgetting and privacy concerns. However, the generated images lack semantic precision and exhibit out-of-distribution characteristics, resulting in inaccurate masks that further degrade the segmentation performance. To tackle these challenges, we propose \mymodel{}, a novel framework featuring a diffusion-based generative replay module that generates semantically accurate images with more reliable masks guided by different instructions (\textit{e.g., text prompts or edge maps}).
Specifically, \mymodel{} introduces a dual-generator paradigm, which focuses on generating old class images that align with the distribution of downstream datasets while preserving the structure and layout of the original images, enabling more precise masks. To adapt to the novel visual concepts of newly added classes continuously, we incorporate class-wise token embedding when updating the dual-generator. Moreover, we assign adequate pseudo-labels of old classes to the background pixels in the new step images, further mitigating the forgetting of previously learned knowledge. Through comprehensive experiments, our method demonstrates competitive performance across mainstream benchmarks, striking a better balance between the performance of old and novel classes.

\end{abstract}

\section{Introduction}

Deep learning has revolutionized various domains by leveraging large pre-collected datasets for training and achieving human-level performance~\cite{floridi2020gpt,kirillov2023segment, radford2021learning, ramesh2022hierarchical}. However, real-world scenarios often involve data streams with storage and privacy constraints, requiring Class-Incremental Learning (CIL) for incremental model updates with new class instances~\cite{shin2017continual,li2017learning,jung2016less,zhou2023deep}. CIL typically faces the challenge of \textit{catastrophic forgetting}, which results in the degradation of performance for old classes as new classes are learned. This issue is more severe in complex tasks like class incremental semantic segmentation (CISS), as recent works emphasize~\cite{cermelli2020modeling,douillard2021plop,cha2021ssul}.

Most existing CISS methods employ the regularization-based methods~\cite{cermelli2020modeling, cha2021ssul, oh2022alife, Baek2022Decomposed}, which aim to alleviate the aforementioned challenges through calibrated loss functions.
In addition, replay-based approaches~\cite{cha2021ssul,oh2022alife,Baek2022Decomposed} offer a complementary solution by replaying a small number of samples from previous steps to retain old class knowledge. This technique is commonly referred to as \textit{memory replay} and serves to mitigate catastrophic forgetting.
%
%
Although demonstrated to be effective, memory replay is limited by privacy concerns~(i.e., the access restrictions of previous step images~\cite{chamikara2018efficient, de2021continual}) 
and the collected samples may not fully capture the diverse semantics within each class. 
To alleviate these problems, \textit{generative replay}-based method~\cite{maracani2021recall} replays old class images produced from a Generative Adversarial Network~(GAN)~\cite{goodfellow2020generative} pre-trained on ImageNet~\cite{deng2009imagenet}. Specifically, this method generate replayed images and use the previous step's model to provide corresponding pseudo-labels. Consequently, the quality of the generated images and pseudo-labels plays a crucial role in CISS performance. 

\begin{figure}[t!]
  \centering
  \includegraphics[width=\linewidth]{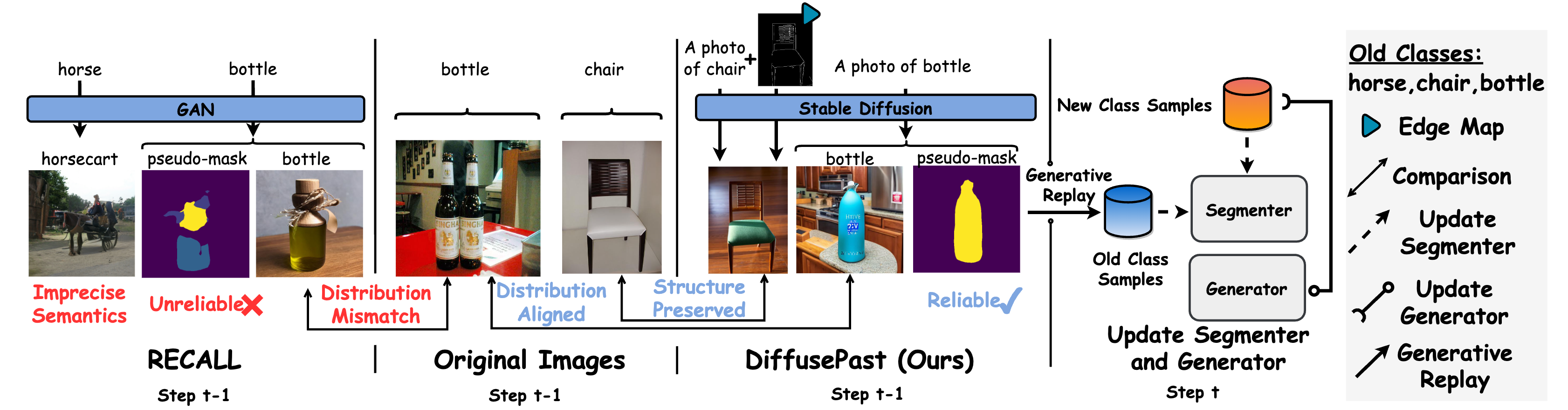}
  \caption{GAN-based generative replay method~(RECALL~\cite{maracani2021recall}) generates inaccurate results; Our \mymodel{}, leveraging Stable Diffusion~\cite{rombach2022high}, generates semantic accurate images of previous steps guided by \textit{text prompts} and \textit{edge maps}. The generated images are distribution-aligned and structure-preserved, which further results in improved masks and benefits CISS task.}
  \label{fig:intro}
\end{figure}

However, this GAN-based method encounters two issues that lead to a decline in the quality of generated images and pseudo-labels, as shown in Figure~\ref{fig:intro}.
Firstly, it generates \textbf{semantically inaccurate} images (e.g., producing a horsecart instead of a horse). This issue arises from an out-of-vocabulary problem, where the horse category is not present in the ImageNet dataset used to train the GAN.
Secondly, even when semantically accurate images are generated, there can be a \textbf{distribution mismatch} between the generated images and the original training data (e.g., generating images with different styles) due to the frozen generator as depicted in~\cite{maracani2021recall}. Consequently, this distribution gap can lead to imprecise pseudo-labels. These two issues, affecting the quality of generated images and pseudo-labels, ultimately contribute to a decline in the overall performance of CISS.

To address these problems, this paper introduces a novel framework named \mymodel{}, which integrates a diffusion-based generative replay module. 
This module leverages the capabilities of pre-trained Stable Diffusion~\cite{rombach2022high} to generate semantically accurate images for replay. 
To enhance the precision of pseudo-labels, we propose a novel dual-generator paradigm consisting of \textit{Distribution-aligned Diffusion Replay} (DDR) and \textit{Structure-preserved Diffusion Replay} (SDR), as shown in Figure~\ref{fig:intro}. Specifically, DDR focuses on generating images that share a similar distribution with the old class training data, improving the reliability of the generated pseudo masks. On the other hand, SDR aims to preserve the semantic structure of the original images from the previous step. For example, it generates new chair images with the same structure (layout) as the original chair images, allowing the use of the original mask annotations as pseudo-labels for the newly generated images. 

Additionally, to adapt to new classes' novel visual concepts during incremental steps, we plug in the class-wise token embedding when updating the dual-generator, thereby preserving knowledge in the pre-trained model and mitigating forgetting by avoiding entire model weights update.
Moreover, to maximize the utilization of old class information, we assign pseudo-labels to the background of new step images for updating segmentation model. This strategy improves the discrimination between old and novel classes, ultimately resulting in a more accurate segmentation model. 
We validate the effectiveness of \mymodel{} through extensive experiments on standard CISS benchmarks~\cite{oh2022alife}.
In summary, the main contributions of our work are as follows:
\begin{itemize}[leftmargin=*]
    \item We introduce the \mymodel{} framework for CISS, which incorporates a diffusion-based generative replay module for generating and replaying semantically accurate samples. This dual-generator paradigm generates distribution-aligned and structure-preserved images, resulting in improved mask annotations and enhanced segmentation performance.
    \item We incorporate class-wise token embedding for continuous adaptation of dual-generator to novel visual concepts of newly added classes, and background pseudo-labeling for updating the segmentation model, which facilitates learning new knowledge while alleviating forgetting old knowledge.
    \item Extensive experiments demonstrate the effectiveness of our method across mainstream benchmarks, showcasing its ability to achieve a better trade-off between the old and novel classes' performance.
    
\end{itemize}


\section{Related work}

\paragraph{Class incremental semantic segmentation}
Most CISS methods fall into two categories: regularization-based methods~\cite{cermelli2020modeling,douillard2021plop,oh2022alife,Baek2022Decomposed,zhang2022mining,zhao2022rbc,phan2022class,michieli2021continual} and replay-based methods~\cite{cha2021ssul, oh2022alife}. Regularization-based methods aim to preserve old class knowledge using calibrated loss functions, mitigating background shift~(i.e., old class pixels in the current step images are labeled as background) and forgetting problems. For example, ALIFE~\cite{oh2022alife} improves the commonly used calibrated cross-entropy and calibrated knowledge distillation losses~\cite{cermelli2020modeling} by introducing the ALI loss, which better preserves old class semantics in the background area, while PLOP~\cite{douillard2021plop} explicitly uses pseudo-labels from the old model's predictions to address background shift. Our method combines the advantages of these approaches to incrementally update the segmentation model through both calibration and pseudo-labeling.
Replay-based methods~\cite{cha2021ssul, Baek2022Decomposed,yan2021framework,zhu2023continual} maintain a small set of samples for each old class to mitigate catastrophic forgetting and improve generalization. However, the classical memory-based approach raises privacy concerns~\cite{chamikara2018efficient, de2021continual} and may lack diversity. On the other hand, the generative replay technique, as presented in RECALL~\cite{maracani2021recall}, generates replayed samples using a pre-trained GAN. However, the generated images suffer from semantic imprecision and encounter out-of-distribution issues, leading to inferior mask annotations and overall performance degradation. As an alternative to using real images, ALIFE~\cite{oh2022alife} proposes replaying class-wise feature embeddings, but it requires additional time-consuming steps to compensate for feature drift when new classes are introduced. Our approach differs from the previous methods in two ways:(1) We generate precise and diverse images directly from text prompts, surpassing ALIFE in obtaining richer class-wise semantics without additional modules or time-consuming feature drift compensation; (2) We continually update the generator by only optimize and store a set of class-wise embeddings, offering greater flexibility compared to the frozen generators in RECALL~\cite{maracani2021recall}.

\paragraph{Synthetic data with generative models}

Synthetic data provides a practical solution for generating large, labeled datasets in vision tasks. These tasks include image classification~\cite{he2022synthetic,azizi2023synthetic}, object detection~\cite{wu2022synthetic,ni2022imaginarynet,ge2022dall}, semantic segmentation~\cite{baranchuk2021label,chen2019learning,li2022bigdatasetgan}. GANs-based methods~\cite{brock2018large, gowal2021improving, li2022bigdatasetgan} have been widely employed in this area.
Recently, there has been increasing interest in using large-scale text-to-image diffusion models such as DALL-E 2~\cite{ramesh2022hierarchical}, Imagen~\cite{saharia2022photorealistic}, and Stable Diffusion~\cite{rombach2022high}). Some works~\cite{li2023guiding, wu2023diffumask} have explored the use of diffusion models to enhance training data for semantic segmentation tasks by leveraging cross-attention maps and fine-tuning. However, these approaches still face challenges related to sequential forgetting~\cite{smith2023continual}, which motivates the need for our work.
\vspace{-0.2cm}
 
\section{Method}

 \vspace{-0.1cm}

\subsection{Problem definition}

During the process of training a Continual Image Semantic Segmentation (CISS) model, we utilize a series of training datasets, denoted as $\mathcal{D}_{t}$, where $t \in \{1, ..., T\}$ refers to a specific incremental step and $T$ represents the total number of steps. For each incremental step $t$, $\mathcal{D}_{t}$ consists of pairs comprising an image $\bm{x}_t \in \mathcal{X}$ and its corresponding ground-truth masks $\bm{y}_t \in \mathcal{Y}_t$. For any sample pairs, $\bm{x}_t = \{\bm{x}_{t,i} \}_{i=1}^N  $ and $\bm{y}_t = \{\bm{y}_{t,i} \}_{i=1}^N$ represent the $N$ pixels of an image and its corresponding mask, respectively.
The label space $\mathcal{Y}_t = \{c_b\} \cup \mathcal{C}_t$ consists of the background class $c_b$ and a set of novel classes $\mathcal{C}_t$, indicating that the masks are labeled only for pixels belonging to categories in $\mathcal{C}_t$, while the remaining pixels are simply labeled as the background $c_b$. 
As a result, the background area of an image $\bm{x}_t$ comprises a collection of pixels denoted by $\mathcal{A}^t_b = \{i | \bm{y}_{t,i} = c_b \}$. The pixels may actually represent objects from the old classes $\mathcal{C}_{1:{t-1}}$, the future classes $\mathcal{C}_{t+1:T}$, or the authentic background itself. In accordance with the prevalent assumption in CISS literature~\cite{Baek2022Decomposed,cermelli2020modeling,cha2021ssul,oh2022alife}, the class sets for each step are considered to be disjoint~(i.e., $\mathcal{C}_{1:t-1} \cap \mathcal{C}_{t} = \emptyset$).

The goal of CISS is to develop a model $f^t_{\bm{\theta}}$ that can identify all previously learned classes. Specifically, at step $t$, $f^t_{\bm{\theta}}$ determines if a pixel is related to any of the known classes: $f^t_{\bm{\theta}} : \mathcal{X} \rightarrow \mathbb{R}^{ N \times | \mathcal{Y}_{1:t} | } $, where $\mathcal{Y}_{1:t} = \{c_b\} \cup \mathcal{C}_{1:t} $. Following previous works~\cite{oh2022alife, cha2021ssul}, $f^t_{\bm{\theta}}$ is constructed as a fully-convolutional network. It consists of a convolutional feature extraction module and a classifier for classifying the classes in $\mathcal{Y}_{1:t}$.
After finishing $t$ learning steps, the prediction for pixel $i$ is allocated to the possible categories encountered so far, denoted as $\hat{\bm y}_{t,i}=\arg\max_{c\in\mathcal{Y}_{1:t}} f_{{\bm{\theta}},c}^t(\bm x_{t,i})$.

\subsection{Overview of the \mymodel{}}

\begin{figure}[t!]
  \centering
  \includegraphics[width=1.0\linewidth]{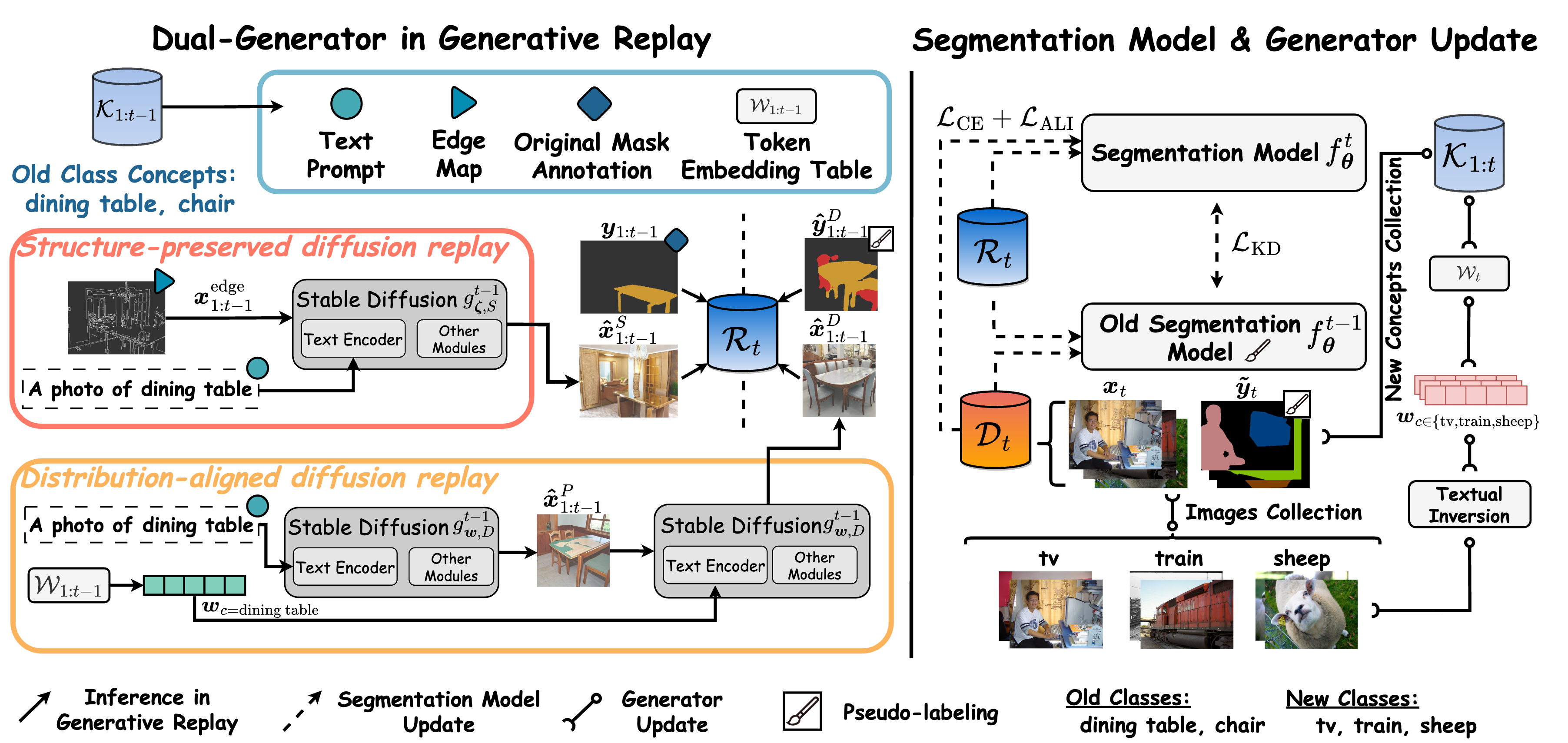}
  \caption{
  \mymodel{} involves two stages:  Inference stage including (1) \textbf{Generative replay} that generates images with reliable masks for replay via dual-generator $g^{t-1}_{\{\bm{\zeta}, \bm{w}\}}$. Training stage including (2) \textbf{Segmentation model update} that updates the segmentation model using the new data $\mathcal{D}_t$, replayed samples $\mathcal{R}_t$, and distilled knowledge from old segmentation model $f^{t-1}_{\bm{\theta}}$; and (3) \textbf{Generator update} that updates the dual-generator as $g^{t}_{\{\bm{\zeta}, \bm{w}\}}$ by collecting and learning novel visual concept from $\mathcal{D}_t$. The old model $f^{t-1}_{\bm{\theta}}$ also provides labels for images generated by the distribution-aligned diffusion replay module and assigns labels to background pixels in images of new dataset $\mathcal{D}_t$.
  }
  \label{fig:overview}
\end{figure}

Our objective is to find a novel and continuous way for replaying images with reliable masks that serve as knowledge of old classes, enabling the update of the segmentation model in a privacy-preserving way and mitigating the problem of catastrophic forgetting.
To achieve these objectives, we introduce \mymodel{}, a generative replay-based framework. The learning process of \mymodel{} involves two stages as shown in Figure~\ref{fig:overview}. 
The pseudo-code of the whole learning procedure is summarized in the supplementary material.
Next, we will provide a detailed description of the two stages:
 \vspace{-0.2cm}
\paragraph{Generative replay.}
Generative replay aims to generate samples $\mathcal{G}_{1:t-1}$ of old classes for updating the current step segmentation model and preventing forgetting. To address semantic imprecision and distribution mismatch in existing generative replay frameworks~\cite{maracani2021recall}, 
we propose a dual-generator $g^{t-1}_{\{\bm{\zeta}, \bm{w}\}} = \{ g^{t-1}_{\bm{\zeta}, {S}}, g^{t-1}_{\bm{w}, {D}} \}$ that leverages pre-trained text-to-image generation models to generate accurate semantic images with text prompt as well as other concepts~(e.g., edge map or class-wise token embedding) of old classes $c\in \mathcal{C}_{1:t-1}$.
Specifically, the \textbf{S}tructure-preserved generator $g^{t-1}_{\bm{\zeta}, {S}}$ aims to generate images $\bm{\hat{x}}_{1:t-1}^{S}$ that preserve the semantic layout, utilizing the text prompt and the structure indicator~(i.e., edge map $\bm{x}_{1:t-1}^{edge}$ derived from the old step images $\bm{x}_{1:t-1}$), while corresponding masks are obtained by reusing the original masks $\bm{y}_{1:t-1}$. The \textbf{D}istribution-aligned generator $g^{t-1}_{\bm{w}, {D}}$ aims to generate images $\bm{\hat{x}}_{1:t-1}^{D}$ that lie in the distribution of downstream datasets, utilizing the text prompt and class-specific token embedding $\bm{w}_c \in \mathcal{W}_{1:t-1}$  while corresponding masks $\bm{\hat{y}}_{1:t-1}^D$ is obtained by pseudo-labeling from old segmentation model $f^{t-1}_{\bm{\theta}}$. The improved images and more reliable masks together benefit the current step segmentation learning~(See Section~\ref{sec:SD} for detailed information on the two generators).
Lastly, the generated samples $\mathcal{G}_{1:t-1}$, consisting of pairs of images and masks $\{ \bm{\hat{x}}_{1:t-1}^{S},\bm{{y}}_{1:t-1} ; \bm{\hat{x}}_{1:t-1}^{D},\bm{\hat{y}}_{1:t-1}^D \}$, are merged to form the current step replay data $\mathcal{R}_t$. 
 \vspace{-0.2cm}
\paragraph{Segmentation model and generator update.}

The replayed samples $\mathcal{R}_t$ of old classes generated in the first stage, and the training set $\mathcal{D}_t$ can be directly used to update the current step's segmentation model $f^t_{\bm{\theta}}$. Additionally, these data can be fed into the old segmentation model $f^{t-1}_{\bm{\theta}}$, whose output logits can be viewed as distilled knowledge and served as another supervision signal for $f^t_{\bm{\theta}}$~(See Section~\ref{sec:loss} for detailed information on the loss functions).
The dual-generator is updated to gain an understanding of novel visual concepts of new classes while preserving the abundant semantic space of the pre-trained generator.
Specifically, for structure-preserved generator $g^{t-1}_{\bm{\zeta}, {S}}$, we update at the \textit{data-level} by collecting the edge maps and the masks $\{ \bm{x}^{\text{edge}}_t, \bm{y}_t \}$ based on the current step dataset $\mathcal{D}_t$. 
For distribution-aligned generator $g^{t-1}_{\bm{w}, {D}}$, we update at the \textit{embedding-level} by learning token embeddings $\bm{w}_c$ that encode the visual concept of each new class $c\in \mathcal{C}_{t}$. 
The above collected concept including the edge map, original masks, class-wise token embedding table, as well as text prompt of new classes, denoted by $\{ \bm{x}_{t}^{edge}, \bm{y}_{t}, \mathcal{W}_{t}, \bm{p}_{t}\}$, respectively, are stored and updated from $\mathcal{K}_{1:t-1}$ to  $\mathcal{K}_{1:t}$, and utilized for inference stage in the next step.

 \vspace{-0.2cm}
\subsection{Dreaming past with diffusion}
\label{sec:SD}
We built a dual-generator upon Stable Diffusion~\cite{rombach2022high} pre-trained on large-scale image-text pairs, to generate semantically accurate images.
Based on this capability, we develop two generators namely the \textit{Structure-preserved Diffusion Replay} and the \textit{Distribution-aligned Diffusion Replay}.

 \vspace{-0.2cm}
\paragraph{{Structure-preserved} diffusion replay.}

To generate images that maintain the semantic structure and layout of the original images $\bm{x}_{1:t-1}$, an intuitive approach is to utilize their masks $\bm{y}_{1:t-1}$ as a guide for image generation. While some existing generation models have demonstrated the ability to convert mask annotations into images~\cite{wang2022semantic}, the inconsistencies in label semantics pose challenges when transferring pre-trained models to downstream datasets.

As a remedy, we utilize the collected edge maps $\bm{x}^{\text{edge}}_{1:t-1}$~(produced from old class images $\bm{x}_{1:t-1}$ by Canny edge detector~\cite{canny1986computational}), and text prompts $\bm{p}_{1:t-1}$ of the old classes as conditions to synthesize images. Formally, given diffusion time step $\tau \in [0,1]$, initial random noise $\bm{\hat{x}}_{\tau = 1}^{S} \sim \mathcal{N}(0,1)$, fixed variance schedule $\beta_\tau$, and conditions $\bm{c} = \{ \bm{p}_{1:t-1}, \bm{x}^{\text{edge}}_{1:t-1} \}$, the structure-preserved old class images $\bm{\hat{x}}_{1:t-1,\tau = 0}^{S}$ can be generated by iteratively applying the reverse diffusion process~(e.g., DDIM~\cite{song2020denoising}):

\begin{equation}
\bm{\hat{x}}_{\tau-1}^{S} = 
\frac{\sqrt{\alpha_{\tau-1}}\bm{x}_{t} - \sqrt{\alpha_{\tau-1}(1-\alpha_{\tau})}\epsilon_{\bm{\phi}}(\bm{\hat{x}}_{\tau}^S,\tau, \bm{c})}{\sqrt{\alpha_{\tau-1}\alpha_{\tau}}} +  
\sqrt{1 - \alpha_{\tau-1}} {\epsilon}_{\bm{\phi}}(\bm{\hat{x}}^{S}_{\tau}, \tau, \bm{c})
\label{eq:reverse}
\end{equation}
where $\alpha_\tau := \prod_{s=0}^{\tau} {(1-\beta_s)}$, $\epsilon_{\bm{\phi}}$ is the pre-trained denoising network. 
In practice, we utilize ControlNet~\cite{zhang2023adding} equipped with Stable Diffusion as the pre-trained denoising network, which supports the canny edge map as the conditions for image generation.
Since synthesized images preserve the spatial consistency of original images~(See qualitative results in Figure~\ref{fig:quali_generate}), thus it is reasonable to pair it with the original masks and construct the samples $\mathcal{G}^S_{1:t-1} = \{ \bm{\tilde{x}}^S_{1:t-1}, \bm{y}_{1:t-1} \}$ for replay. Lastly, we add the edge map $\bm{x}_{t}^{edge}$, current step masks $\bm{y}_{t}$, as well as text prompt $\bm{p}_{t}$ to $\mathcal{W}_{1:t}$ for next step inference.

 \vspace{-0.2cm}
\paragraph{Distribution-aligned diffusion replay.}

To generate images that preserve the diversity and align with the distribution of the downstream datasets, we follow the image-to-image translation process in~\cite{meng2021sdedit}, dividing the generation process into two sub-steps. Firstly, we generate diverse images $\bm{\hat{x}}_{1:t-1}^{P}$ based on old class text prompts $\bm{p}_{1:t-1}$ from scratch, and then we perform the distribution transfer on $\bm{\hat{x}}_{1:t-1}^{P}$ via class-wise token embeddings $\mathcal{W}_{1:t-1}$ of the old classes $c \in \mathcal{C}_{1:t-1}$. Formally, we first add the noise $\bm{\epsilon} \sim \mathcal{N}(0,1)$ to obtain the noised images of $\bm{\hat{x}}_{1:t-1}^{P}$ by:
\begin{align}
    \bm{\hat{x}}^P_{\tau = 1} = \sqrt{{\alpha}_{\tau}} \bm{\hat{x}}^P_{\tau = 0} + \sqrt{1 - {\alpha}_{\tau}} \bm{\epsilon}
\end{align}
where $\tau$ and $\alpha$ share the same definition in Eq.\ref{eq:reverse}. Then, we perform the reverse diffusion $\bm{\hat{x}}_{\tau=1}^{P} \rightarrow \bm{\hat{x}}_{\tau=0}^{D}$ by iterating Eq.\ref{eq:reverse} with conditions $\bm{c} = \mathcal{W}_{1:t-1}$. The labels for generated images will be given by the old segmentation model via pseudo-labeling $\bm{\hat{y}}^{D}_{1:t-1,i}=\arg\max_{c\in\mathcal{Y}_{1:t-1}} f_{\bm{\theta},c}^{t-1}(\bm{\hat{x}}^D_{1:t-1,i})$ for each pixel $i$. Finally, we construct the samples $\mathcal{G}^D_{1:t-1} = \{ \bm{\hat{x}}^D_{1:t-1}, \bm{\hat{y}}^D_{1:t-1} \}$ for replay.


To update the generator, we learn the token embeddings $\bm{w}_c \in\mathcal{W}_t$ of each new class $c\in \mathcal{C}_t$ leveraging the technique of Textural Inversion~\cite{gal2022image}. Formally, we add $|\mathcal{C}_t|$ new tokens to the text vocabulary of Stable Diffusion for each new class and fine-tune these embeddings $\bm{w}_c$ using the diffusion loss:
\begin{equation}
\begin{split}
    \mathcal{L}_\text{token} & (\bm{w}_c)_{c\in \mathcal{C}_t} = 
    \mathbb{E} \left[ \| \bm{\epsilon} - \epsilon_{\phi} ( \bm{x}_{c, \tau}, \tau ) \|^2 \right]
\end{split}
\label{eq:forward}
\end{equation}

where training image $\bm{x}_{c, \tau = 0}$ is collected for each class $c \in \mathcal{C}_t$ with few-shot~(e.g., 4) samples. Lastly, we add the learned token embeddings to $\mathcal{W}_{1:t}$.
Since the optimization of each class-related token is independent and not disrupting the pretrained weights of Stable Diffusion, it does not face the forgetting problem.


 \vspace{-0.2cm}
\subsection{Training}
\label{sec:loss}

The optimization of the segmentation model $f^{t-1}_{\bm{\theta}}$ based on three losses.
These include the Cross-Entropy (CE) loss and ALI loss~\cite{oh2022alife} for learning from generative replay samples $\mathcal{R}_t$ and current step samples $\mathcal{D}_t$, and knowledge distillation~(KD) loss for learning from the old segmentation model. Nevertheless, the majority of existing works~\cite{cha2021ssul, oh2022alife} use CE loss to learn the discrimination boundary solely based on labels of new classes $c \in \mathcal{C}_t$, overlooking the old class pixels in the background area.
Motivated by~\cite{douillard2021plop}, we aim to improve the discrimination by applying CE loss on both the ground-truth $\mathcal{A}^{t}_\text{new} = \{i | \bm{y}_{t,i} \in \mathcal{C}_t \}$ and pseudo-labeled areas $\mathcal{A}^{t}_\text{old} = \{i | (\bm y_{t,i}=c_b) \wedge (\hat{\bm y}_{t,i}\in\mathcal{C}_{1:t-1}) \wedge (\bm\mu_i >\tau) \}$, where predictions $\hat{\bm y}_{t,i}=\arg\max_{c\in\mathcal{Y}_{1:t-1}} f_{{\bm{\theta}},c}^{t-1}(\bm x_{t,i}) $ are assigned to the background pixels $\mathcal{A}^t_b = \{i | \bm{y}_{t,i} = c_b \}$ only when the predictions correspond to old class objects are confident. Finally, the overall objective is:
\begin{equation} \label{eq:loss}
	\mathcal{L}(i) 
	= \lambda_\mathrm{CE}\mathcal{L}_\mathrm{CE}(i)\mathbbm{1}[i \in \mathcal{A}^{t}_\mathrm{new}\cup \mathcal{A}^{t}_\mathrm{old}]
	+ \lambda_\mathrm{ALI} \mathcal{L}_\mathrm{ALI}(i)\mathbbm{1}[i \in \mathcal{A}^{t}_{b}]
	+ \lambda_\mathrm{KD} \mathcal{L}_\mathrm{KD}(i),
\end{equation}
where$\lambda_\mathrm{CE}$, $\lambda_\mathrm{ALI}$ and $\lambda_\mathrm{KD}$ are balance parameters. $\mathcal{L}_{\mathrm{KD}}$ calculates the dot-product distance between the old and new segmentation model's output probabilities, while $\mathcal{L}_\mathrm{ALI}$ is the distance between the maximum logit value and the weighted average of old class logits~\cite{oh2022alife}.
 \vspace{-0.1cm}
\section{Experiments}
 \vspace{-0.1cm}
\subsection{Experimental setting}
\paragraph{Datasets and evaluation.} We start by evaluating our method on the PASCAL-VOC 2012 \cite{everingham2010pascal} and ADE20K \cite{zhou2017scene} datasets. The Pascal-VOC 2012 dataset covers 20 object and background classes. The ADE20K dataset contains 150 object and stuff classes~(See supplementary material for more details of datasets).
We investigated a variety of incremental learning scenarios for each dataset following the common practice~\cite{cermelli2020modeling,cha2021ssul, oh2022alife}. The incremental scenario is denoted by $N_\text{b}$-$N_\text{n}$ ($T$ steps), where $N_b$ and $N_n$ represent the number of base and novel classes to be trained, respectively. 
For instance, 
an incremental scenario of 15-1 (6 steps) would involve learning 15 base classes first, followed by sequentially adding one novel class, requiring a total of 6 training steps. 
Following~\cite{oh2022alife,cha2021ssul}, we focus on the \textit{overlapped} setting, where unlabeled regions could contain either previous or future categories, as it is more practical in real-world applications.
In line with \cite{cermelli2020modeling, douillard2021plop, michieli2021continual, cha2021ssul}, we evaluate our method and all baselines using the $\text{mIoU}_{\text{b}}$, $\text{mIoU}_{\text{n}}$, and $\text{mIoU}_{\text{all}}$ scores, which represent the average intersection-over-union (IoU) scores for base, novel, and all categories, respectively. We also provide the harmonic mean (hIoU) of $\text{mIoU}_{\text{b}}$ and $\text{mIoU}_{\text{n}}$ scores, which is less affected by the imbalance between base and novel classes.

\textbf{Implementation details.}
We use DeepLab-V3 with ImageNet pre-trained ResNet-101 as the segmentation model, following \cite{cermelli2020modeling, cha2021ssul, douillard2021plop, oh2022alife}. 
We employ the Stable Diffusion~\cite{rombach2022high} to generate images conditioned on text prompts, 
and utilize the ControlNet~\cite{zhang2023adding} to control the Stable Diffusion, allowing for edge map conditions.
We optimize the network using SGD with an initial learning rate of 1e-2 for the base stage and 1e-3 for the incremental stages. During the base stage ($t=1$), DeepLab-V3 is trained for 30 epochs on PASCAL VOC and 60 epochs on ADE20K. For each subsequent incremental stage ($t>1$), the number of epochs for PASCAL VOC is determined by cross-validation, while ADE20K is trained for a fixed 60 epochs, as in~\cite{oh2022alife}. The batch size is set to 24 on both datasets. 
We use the poly schedule for adjusting the learning rate. We follow the learning rate schedule, data augmentation, and output stride outlined in \cite{oh2022alife} for all experiments.
The replayed sample size is set as 100 for Pascal VOC and 300 for ADE20K, following~\cite{cha2021ssul}.
The experiments were conducted using PyTorch 1.13.1~\cite{paszke2019pytorch} on two NVIDIA A100 GPUs.
We report IoU scores averaged over 3 runs.
A detailed description of hyperparameter settings can be found in the supplementary material.

\vspace{-6pt}
\subsection{Experimental results on benchmark datasets}

\begin{table}[htbp]
  \setlength{\tabcolsep}{0.21em}
  \centering
  \scriptsize
  \caption{Experimental results on PASCAL-VOC~\cite{everingham2010pascal}. ALIFE-M uses feature replay, while RECALL uses generative replay. Note that RECALL~\cite{maracani2021recall} generates 500 samples for each old class.
  Numbers in bold are the best performance, while underlined ones are the second best. 
  Numbers for other methods are taken from ALIFE~\cite{oh2022alife}.}
  \label{tab:sota_voc}
  \begin{tabular}{L{1.85cm} C{0.85cm}C{0.85cm}C{0.85cm}C{0.85cm}
  C{0.85cm}C{0.85cm}C{0.85cm}C{0.85cm}
  C{0.85cm}C{0.85cm}C{0.85cm}C{0.85cm}}
      \toprule
      \multirow{2.5}{*}{Methods}
      & \multicolumn{4}{c}{\textbf{19-1~(2 steps)}}
      & \multicolumn{4}{c}{\textbf{15-5~(2 steps)}}
      & \multicolumn{4}{c}{\textbf{15-1~(6 steps)}} 
      \\

      \cmidrule(lr){2-5} \cmidrule(lr){6-9} \cmidrule(lr){10-13}
      & 0-19 & 20 & mIoU & hIoU
      & 0-15 & 16-20 & mIoU & hIoU
      & 0-15 & 16-20 & mIoU & hIoU 
      \\
      
      \midrule
	  EWC~\cite{kirkpatrick2017overcoming}
      & 26.90 & 14.00 & 26.30 & 18.42
      & 24.30 & 35.50 & 27.10 & 28.85
      &~~0.30 &~~4.30 &~~1.30 &~~0.56
      \\ 

	  LwF-MC~\cite{li2017learning}
      & 64.40 & 13.30 & 61.90 & 22.05
      & 58.10 & 35.00 & 52.30 & 43.68
      &~~6.40 &~~8.40 &~~6.90 &~~7.26
      \\ 
      
	  ILT~\cite{michieli2019incremental}
      & 67.75 & 10.88 & 65.05 & 18.75
      & 67.08 & 39.23 & 60.45 & 49.51
      &~~8.75 &~~7.99 &~~8.56 &~~8.35 
      \\ 
      
      MiB~\cite{cermelli2020modeling}
      & 70.42 & 17.70 & 67.91 & 28.25
      & 76.68 & 49.03 & 70.09 & 59.81
      & 37.98 & 12.28 & 31.86 & 18.56 
      \\ 
      
      SDR~\cite{michieli2021continual}
      & 71.30 & 23.40 & 69.00 & 35.24 
      & 76.30 & {50.20} & 70.10 & 60.56
      & 47.30 & 14.70 & 39.50 & 22.43 
      \\ 

      PLOP~\cite{douillard2021plop}
      & {75.89} & 34.90 & {73.94} & {47.81}
      & 76.37 & 49.55 & 69.98 & 60.10
      & {64.51} & {19.93} & {53.90} & {30.45} 
      \\ 

	  SSUL~\cite{cha2021ssul}
       & {77.73} & 29.68 & \underline{75.44} & 42.96
      & {77.82} & 50.10 & {71.22} & {60.96}
      & {77.31} & {36.59} & \textbf{67.61} & \textbf{49.67}
      \\ 
         
      ALIFE~\cite{oh2022alife}
      & 76.61 & {49.36} & 75.31 & \underline{60.03}
      & {77.18} & {52.52} & \underline{71.31} & \underline{62.50} 
      & {64.44} & {34.91} & {57.41} & {45.29} 
      \\

      \multirow{1.75}{*}{\textbf{{\myloss{}}}}
      & {77.59} & {54.00} & \textbf{75.51} & \textbf{63.35}
      & 78.05   & 53.13    & \textbf{72.10}& \textbf{63.22}
      & {63.94} & {36.64} & \underline{57.44} & \underline{46.58} 
      \\ [-0.25em]
      & {\tiny{$\pm$0.01}}
      & {\tiny{$\pm$0.00}}
      & {\tiny{$\pm$0.01}}
      & {\tiny{$\pm$0.00}}
      & {\tiny{$\pm$0.02}}
      & {\tiny{$\pm$0.12}}
      & {\tiny{$\pm$0.03}}
      & {\tiny{$\pm$0.10}}
      & {\tiny{$\pm$0.69}}
      & {\tiny{$\pm$0.04}}
      & {\tiny{$\pm$0.53}}
      & {\tiny{$\pm$0.21}}
      \\
      \midrule
        ALIFE-M~\cite{oh2022alife}
      & {76.72} & {52.29} & \underline{75.56} & \underline{62.19}
      & {77.66} & {55.27} & \underline{72.33} & \underline{64.57}
      & {66.09} & {38.81} & {59.59} & {48.89}
      \\
         RECALL~\cite{maracani2021recall}
      & {68.10} & {55.30} & {68.60} & {61.04}
      & {67.70} & {54.30} & {65.60} & {60.26} 
      & {67.80} & {50.90} & \underline{64.80} & \underline{58.15} 
      \\ 
       \multirow{1.75}{*}{\textbf{Ours-\mymodel{}}}
        & {76.75} & {57.16} & \textbf{75.82} & \textbf{65.52}
       & 78.47  & 57.36 & \textbf{73.45} & \textbf{66.27}
      & {74.68} & {49.17} & \textbf{68.60} & \textbf{59.30}
      \\ [-0.25em]
      & {\tiny{$\pm$0.04}}
      & {\tiny{$\pm$0.04}}
      & {\tiny{$\pm$0.04}}
      & {\tiny{$\pm$0.04}}
      & {\tiny{$\pm$0.01}}
      & {\tiny{$\pm$0.04}}
      & {\tiny{$\pm$0.01}}
      & {\tiny{$\pm$0.03}}
      & {\tiny{$\pm$0.17}}
      & {\tiny{$\pm$0.25}}
      & {\tiny{$\pm$0.18}}
      & {\tiny{$\pm$0.22}}
      \\
      
      \bottomrule
  \end{tabular}
\end{table}

\begin{figure}[h!]
    \centering
    \subfigure[]{           
        \includegraphics[width=0.31\linewidth]{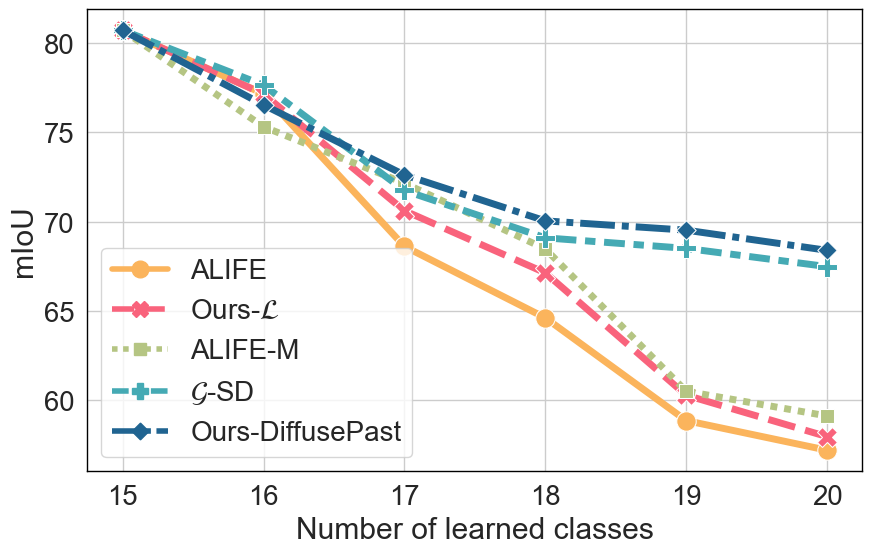} 
        \label{fig:miou_evo}  
    }
    \subfigure[]{           
        \includegraphics[width=0.31\linewidth]{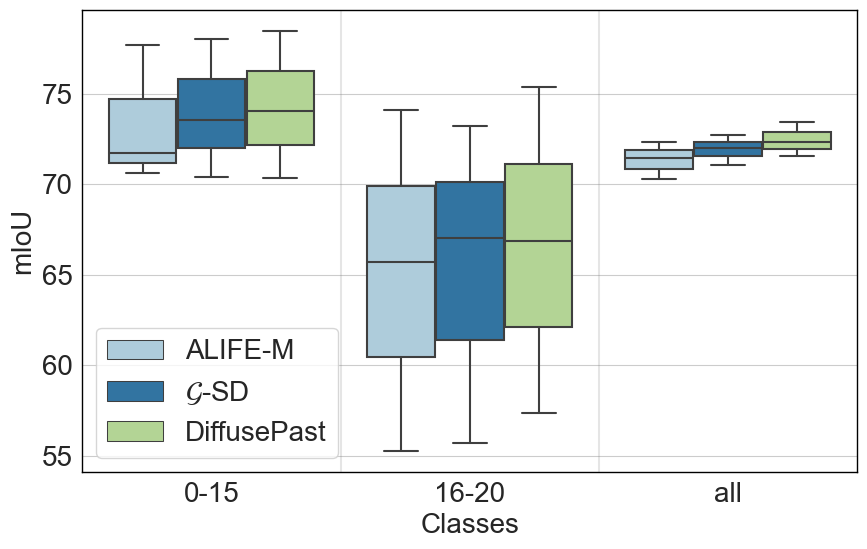} 
        \label{fig:class_ordering}         
    }
    \subfigure[]{           
        \includegraphics[width=0.31\linewidth]{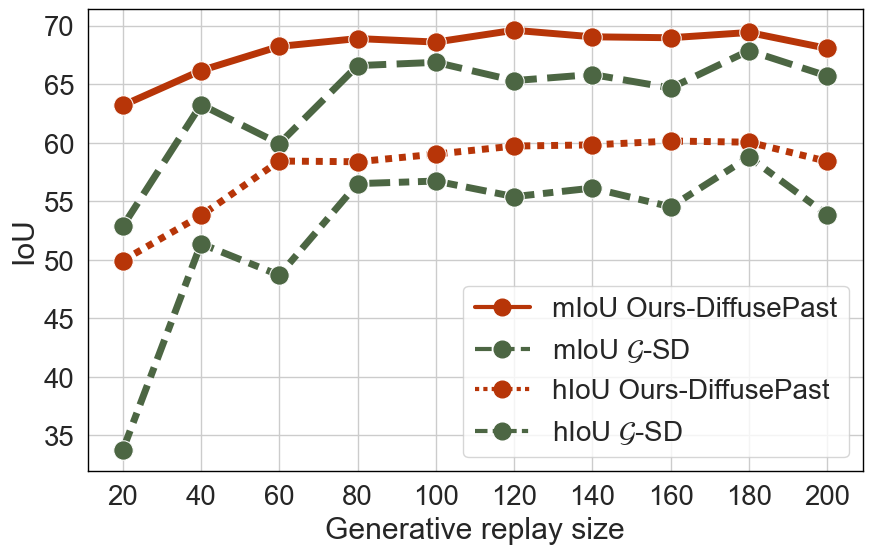} 
        \label{fig:replay_size}         
    }
    \caption{ (a): mIoU evaluation on VOC 15-1, (b): Comparison of mIoU distribution for three different class-orderings on VOC 15-5, (c): mIOU and hIoU on VOC 15-1 with varying replay size.}
    \label{fig:voc_analysis}
\end{figure}

\begin{figure}[h!]
  
  \centering
  \includegraphics[width=\linewidth]{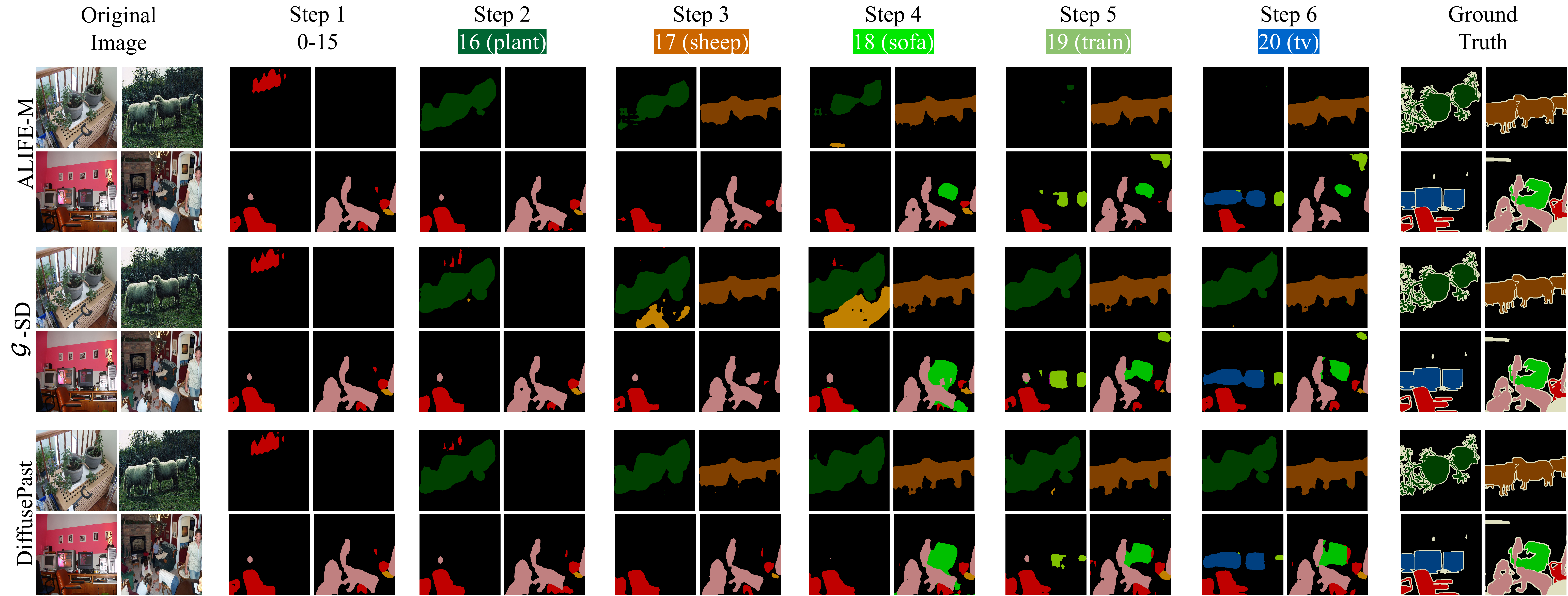}
  \caption{Qualitative results on VOC 15-1 scenario.}
  \label{fig:quali_voc}
\end{figure}

\paragraph{PASCAL VOC.}
We compare our approach in Table~\ref{tab:sota_voc} with state-of-the-art CISS methods, including regularization-based methods~\cite{oh2022alife,cha2021ssul,cermelli2020modeling,douillard2021plop} and replay-based methods that do not use real images~\cite{oh2022alife,maracani2021recall}.
We have the following observations:
(1) The proposed loss, \myloss{}, outperforms all other methods without replay~(especially ALIFE, which also uses ALI loss) in 19-1 and 15-5 scenarios. This highlights the effectiveness of utilizing old class knowledge in the cross-entropy loss. Note that a comparison of SSUL~\cite{cha2021ssul} is not fair due to its use of a saliency detector that allows more classes~(i.e., future objects) for discrimination learning on 15-1.
(2) Replay strategies without real images (ALIFE-M, RECALL, and \mymodel{}) effectively consolidate old class knowledge and enhance novel class learning. This validates the intuition that by involving old pixels, the model learns with more information, resulting in improved discrimination boundaries.
(3) Our proposed \mymodel{}, utilizing diffusion-based generative replay, outperforms other replay-based methods with significant improvements~(the largest mIoU gains of 3.80\% on 15-1 and hIoU gains of 3.33\% on 19-1), and achieve substantial gains over \myloss{}~(the largest mIoU and hIoU gains of 11.16\% and 12.72\% on 15-1). These results confirm the effectiveness of our approach. 

To further analyze the more challenging tasks 15-1, we show the mIoU evolution at each incremental step in Figure~\ref{fig:miou_evo}.
The results show that the proposed loss function (\myloss{}) consistently outperforms ALIFE at every step and even matches the performance of ALIFE-M. These results indicate that the properly utilized pseudo-labels are effective for CISS.
Besides, we found that generative replay-based methods (\mymodel{} and $\mathcal{G}$-SD) achieve substantial increases in mIoU when new classes are incrementally learned. Moreover, compared to $\mathcal{G}$-SD, which replays images generated by vanilla Stable Diffusion, \mymodel{} consistently achieves better performance when $t\geq3$. These results validate the effectiveness of the proposed dual-generator and confirm the intuition that improved images and reliable masks benefit the final performance.
We also evaluate the performance robustness of \mymodel{} on different class orderings in VOC 15-5. Figure~\ref{fig:class_ordering} shows that \mymodel{} consistently achieves higher mIoUs than other methods, demonstrating its robustness.

\begin{figure}[h]
  \centering
  \includegraphics[width=0.95\linewidth]{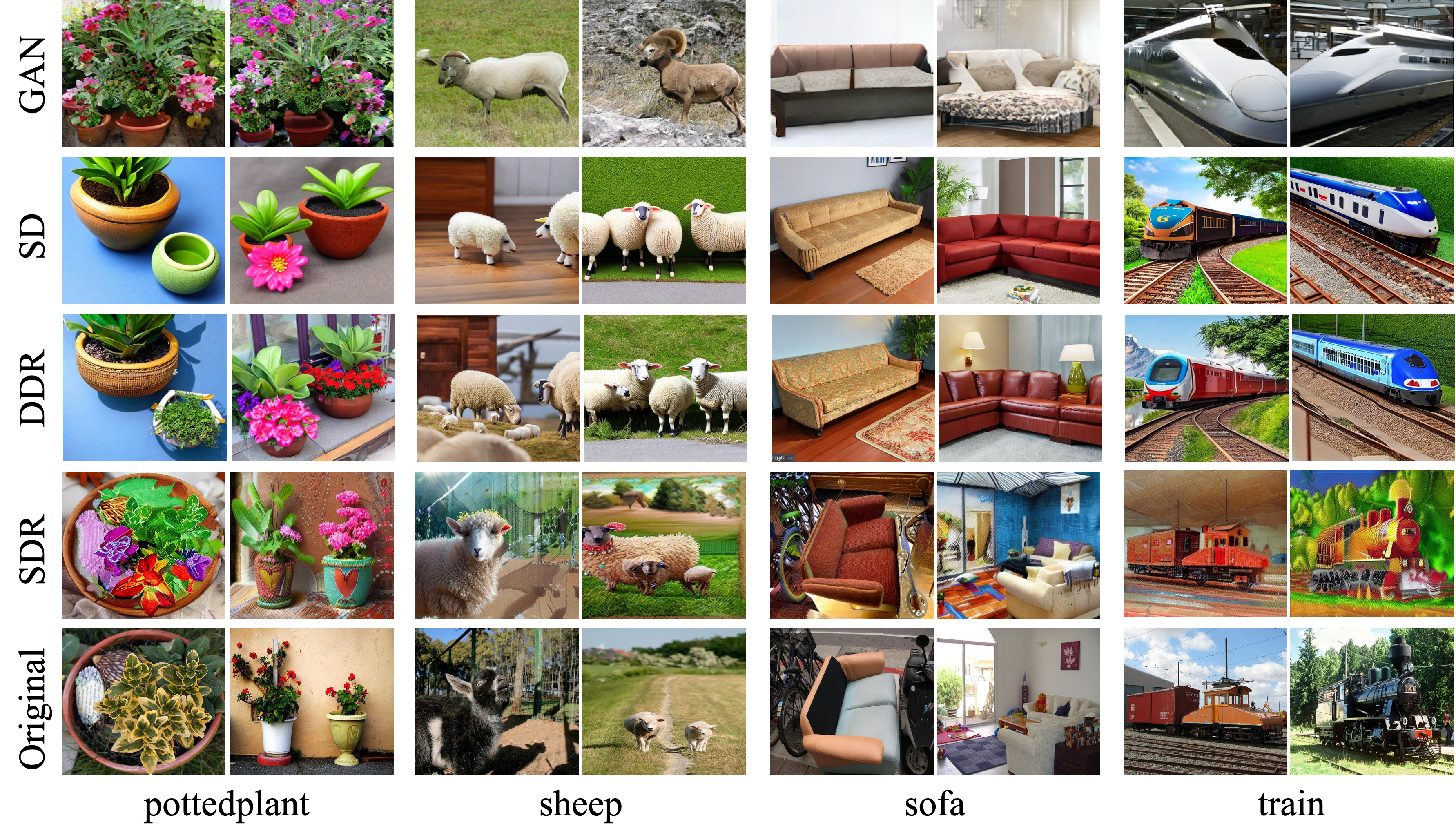}
  \caption{
  Visualization of generated images on VOC 15-1 at last four incremental steps with GAN, Stable-Diffusion (SD), Structure-preserved Diffusion Replay~(SDR), and Distribution-aligned Diffusion Replay~(DDR). {\textit{pottedplant}}, {\textit{sheep}}, \textit{sofa} and {\textit{train}} are generated samples for replay.
  }
  \label{fig:quali_generate}
\end{figure}

\textbf{Qualitative results.} 
Figure~\ref{fig:quali_voc} shows a qualitative analysis of our approach compared to other replay-based methods (ALIFE-M and $\mathcal{G}$-SD), showcasing results for four images per learning step in the VOC 15-1 task. These images contain the newly learned classes \{\textit{plant}, \textit{sheep}, \textit{sofa}, \textit{train}, \textit{tv}\} as well as old classes~\{\textit{person}, \textit{chair}\}. Our \mymodel{} exhibits both less forgetting and better new class learning based on the following observations:
First, unlike ALIFE-M, the generative-replay based methods preserve the knowledge of base step classes (e.g., \textit{person} and \textit{chair}) and intermediate step classes (e.g., \textit{plant}). Further, \mymodel{} retain more old class knowledge~(e.g., \textit{person}) from the base steps compared with $\mathcal{G}$-SD;
Second, we observe that both $\mathcal{G}$-SD and \mymodel{} effectively learn new classes~(e.g., \textit{sheep}). In addition, \mymodel{} effectively mitigates the false-positive predictions~(e.g., predicted \textit{sofa} near the \textit{tv} in Step 5). We also compare the old class images produced by different generative-replay strategies. Figure~\ref{fig:quali_generate} shows that the images generated by distribution-aligned diffusion replay~(DDR) more closely match the original data than GAN and SD, while the structure-preserved diffusion replay~(SDR) generates images with layouts similar to the original images, validating the rationale of reusing the original masks. These improvements validate the effectiveness of the proposed dual-generator and contribute to the overall performance enhancement.


\begin{table}[t!]
  \setlength{\tabcolsep}{0.21em}
  \centering
  \scriptsize
  \caption{Experimental results on ADE20K~\cite{zhou2017scene}. 
  Numbers in bold are the best performance, while underlined ones are the second best. Numbers for other methods are taken from ALIFE~\cite{oh2022alife}.}
  \label{tab:sota_ade}
  \begin{tabular}{L{1.85cm} C{0.85cm}C{0.85cm}C{0.85cm}C{0.85cm}
  C{0.85cm}C{0.85cm}C{0.85cm}C{0.85cm}
  C{0.85cm}C{0.85cm}C{0.85cm}C{0.85cm}}
      \toprule
      \multirow{2.5}{*}{Methods}
      & \multicolumn{4}{c}{\textbf{100-50~(2 steps)}}
      & \multicolumn{4}{c}{\textbf{50-50~(3 steps)}}
      & \multicolumn{4}{c}{\textbf{100-10~(6 steps)}} 
      \\
      
      \cmidrule(lr){2-5} \cmidrule(lr){6-9} \cmidrule(lr){10-13}
      & 0-100 & 101-150 & mIoU & hIoU
      & 0-50 & 51-150 & mIoU & hIoU
      & 0-100 & 101-150 & mIoU & hIoU 
      \\

      \midrule
      ILT~\cite{michieli2019incremental}
      & 18.29 & 14.40 & 17.00 & 16.11
	  &~~3.53 & 12.85 &~~9.70 &~~5.54
      &~~0.08 &~~1.31 &~~0.49 &~~0.15
      \\

      MiB~\cite{cermelli2020modeling}
      & 40.52 & 17.17 & 32.79 & 24.12 
	  & 45.57 & 21.01 & 29.31 & 28.76
	  & 38.21 & 11.12 & 29.24 & 17.23
      \\

      PLOP~\cite{douillard2021plop}
      & {42.10} & {16.22} & {33.53} & {23.42}
      & {48.24} & {21.31} & {30.40} & {29.56}
      & {40.78} & {14.02} & {31.92} & {20.87} 
      \\
          
	  SSUL~\cite{cha2021ssul}
      & {41.28} & {18.02} & {33.58} & {25.09}
      & {48.38} & {20.15} & {29.56} & {28.45}
      & {40.20} & {18.75} & {33.10} & {25.57}
      \\
      
        ALIFE~\cite{oh2022alife}
      & {42.18} & {23.07} & \underline{35.86} & \underline{29.83}
      & {48.98} & {25.69} & \underline{33.56} & \underline{33.70} 
      & {41.02} & {22.76} & \underline{34.98} & \textbf{29.28}
      \\

        \multirow{1.75}{*}{\textbf{\myloss{}}}
      & {42.12} & {24.87} & \textbf{36.41} & \textbf{31.28}
      & {48.78} & {28.92} & \textbf{35.62} & \textbf{36.31} 
      & {41.66} & {22.20} & \textbf{35.21} & \underline{28.96}
      \\ [-0.25em]
      & {\tiny{$\pm$0.00}}
      & {\tiny{$\pm$0.02}}
      & {\tiny{$\pm$0.01}}
      & {\tiny{$\pm$0.02}}
      & {\tiny{$\pm$0.00}}
      & {\tiny{$\pm$0.01}}
      & {\tiny{$\pm$0.01}}
      & {\tiny{$\pm$0.01}}
      & {\tiny{$\pm$0.23}}
      & {\tiny{$\pm$0.16}}
      & {\tiny{$\pm$0.11}}
      & {\tiny{$\pm$0.07}}
      \\
      \midrule
	ALIFE-M~\cite{oh2022alife}
      & {42.28} & {23.58} & \underline{36.09} & \underline{30.28}
      & {48.99} & {26.15} & \underline{33.87} & \underline{34.10} 
      & {41.17} & {23.07} & \underline{35.18} & \underline{29.57}
      \\
        \multirow{1.75}{*}{\textbf{Ours-\mymodel{}}}
        & {42.26} & {24.86} & \textbf{36.50} & \textbf{31.31}
      & {48.93} & {29.13} & \textbf{35.82} & \textbf{36.52} 
      & {41.76} & {23.85} & \textbf{35.83} & \textbf{30.36}
      \\[-0.25em]
      & {\tiny{$\pm$0.02}}
      & {\tiny{$\pm$0.04}}
      & {\tiny{$\pm$0.00}}
      & {\tiny{$\pm$0.02}}
      & {\tiny{$\pm$0.01}}
      & {\tiny{$\pm$0.04}}
      & {\tiny{$\pm$0.02}}
      & {\tiny{$\pm$0.02}}
      & {\tiny{$\pm$0.04}}
      & {\tiny{$\pm$0.18}}
      & {\tiny{$\pm$0.03}}
      & {\tiny{$\pm$0.13}}
      \\
      \bottomrule[1pt]
  \end{tabular}
  \vspace{-0.3cm}
\end{table}

\vspace{-5pt}
\paragraph{ADE20K.}
Table~\ref{tab:sota_ade} presents quantitative results on ADE20K, a densely labeled dataset with rich semantics in the background area at each incremental step. Note that RECALL~\cite{maracani2021recall} is not designed to handle stuff categories, thus results are not available. Similar to the observations in VOC, the results again demonstrate the effectiveness of \myloss{} and Ours-\mymodel{} for densely labeled datasets. The qualitative results on ADE20K can be found in the supplementary material.

\subsection{Discussion}

\setlength\intextsep{0pt}
\begin{wraptable}{r}{0.55\textwidth}
  \centering
  \smallskip\noindent
  \caption{Ablation study for generative replay $\mathcal{G}$ with Stable-Diffusion (SD), Structure-preserved Diffusion Replay~(SDR), and Distribution-aligned Diffusion Replay~(DDR) on VOC 15-5 and ADE20K 50-50. Memory replay-based methods~(SSUL-M, Ours-M) are also included for comparison.
  }
  \label{tab:abla_generator}
   \resizebox{\linewidth}{!}
	{
      \begin{tabular}{l cccc cccc}
      \toprule[1pt]
            \multirow{2.5}{*}{Methods} & 
             \multicolumn{4}{c}{{\textbf{15-5~(2 steps)}}}  &
             \multicolumn{4}{c}{{\textbf{50-50~(3 steps)}}}\\
             \cmidrule(lr){2-5}  \cmidrule(lr){6-9}
               & 0-15 & 16-20  & mIoU & hIoU &
            0-50 & 51-150  & mIoU & hIoU \\
      \midrule[1pt]
        SSUL-M~\cite{cha2021ssul}  & 78.40& 55.80 & 73.02& 65.20 & 49.12 & 20.10 & 29.77 &28.53\\ 
      
        Ours-M  & 77.83 & 55.26 & 72.46 & 64.63 & 48.71 & 29.00 & 35.66 & 36.36\\ 
      \midrule
      $\mathcal{G}-\text{SD}$  & 78.02
        & 55.77 & 72.72 & 65.04 & 48.77	& 28.84 & 35.57 & 36.25 \\
      $\mathcal{G}-\text{SDR}$  &    77.92 & 56.92  & 72.92 & 65.79 & 48.86 & 28.77 & 35.56 & 36.22\\
      $\mathcal{G}-\text{DDR}$  & 78.45  &  56.58& 73.24 & 65.74 & 48.93&29.13 & 35.82 & 36.52 \\
        
    \mymodel{} & 78.47  &  57.36& \textbf{73.45} & \textbf{66.27} & 48.93 & 29.13 & \textbf{35.82} & \textbf{36.52} \\
      \bottomrule[1pt]
    \end{tabular}
  }
  \vspace{5pt}
\end{wraptable}

\paragraph{Ablation study on generator.} 
Table~\ref{tab:abla_generator} presents the effect of each generator in \mymodel{} on VOC 15-5 and ADE20K 50-50.
The last four lines in the tables compare the performance between different generation paradigms. The results show that SDR and DDR perform better than SD on VOC 15-5, while DDR outperforms SD on ADE20K 50-50. This performance degradation of SDR on ADE20K may attribute to the difficulty of generating high-quality images on complex scenes. As a result, our \mymodel{} merges the samples with a ratio of 0.8 for DDR and achieves better performance on VOC 15-5 while using DDR-generated samples directly on ADE20K 50-50.
We also compare the results with memory replay-based methods. The first two lines in the table show that our \mymodel{} outperforms SSUL-M and Ours-M~(\myloss{} equipped with memory-replay) in both scenarios. A plausible reason is that the frozen feature extractor in SSUL-M limits its adaptability to new class knowledge. Another reason may be that our \mymodel{} generates diverse old class samples at each step, providing more knowledge compared to fixed memory replay.

\vspace{-5pt}
\setlength\intextsep{0pt}
\begin{wraptable}{r}{0.55\textwidth}
  \centering
  \smallskip\noindent
  \caption{
    Comparison of IoU scores using different loss terms on VOC 15-5. \textit{Pseudo} indicates that CE is applied on pseudo-labeled regions. \textit{bg} logit indicates that KD is applied on both old class logits and the \textit{bg} logit.
  }
  \label{tab:abla_loss}
  \resizebox{\linewidth}{!}
	{
		\begin{tabular}{ cc  c c C{1.25cm}C{1.25cm}C{1.25cm}C{1.25cm}}
			\toprule[1pt]
			 \multicolumn{2}{c}{CE} &   \multicolumn{1}{c}{KD}  & \multirow{2.5}{*}{$\mathcal{G}$} &\multirow{2.5}{*}{0-15} & \multirow{2.5}{*}{16-20} & \multirow{2.5}{*}{mIoU} & \multirow{2.5}{*}{hIoU}
			\\
			\cmidrule(lr){1-2}\cmidrule(lr){3-3}
   
             \textit{Ground-truth}&\textit{Pseudo} & \textit{bg} logit  & &  & & 
			\\
            \midrule
			 \checkmark &  & \checkmark & & 76.96 & 51.22 & 70.83 & 61.52
			\\
                \checkmark &  & & & 78.00 & 52.93 & \underline{72.03} & \underline{63.07}
			\\
			 \checkmark & \checkmark &   &  & 78.05 & 53.13 & \textbf{72.10} & \textbf{63.22}
			\\
            \midrule
			 \checkmark &  & \checkmark &  \checkmark & 78.13  & 56.40 & 72.96  & 65.51
			\\
                \checkmark &  & & \checkmark & 78.30 & 57.22&\underline{73.28} & \underline{66.12}
			\\
			 \checkmark & \checkmark &  & \checkmark &78.47  & 57.36 & \textbf{73.45} & \textbf{66.27}
                 \\
			\bottomrule[1pt]
		\end{tabular}
	}
  \vspace{2pt}
\end{wraptable}
\paragraph{Ablation study on loss function.}
Table~\ref{tab:abla_loss} presents an ablation analysis of \myloss{}~(Eq.~\ref{eq:loss}).
The first row shows that using the KD on the \textit{bg} logit results in degraded performance. This is because the old segmentation model tends to assign low probabilities to new classes, and distilling such logit will interfere with the learning of new classes.
The second and third rows highlight the effectiveness of the pseudo-labels strategy, which mitigates forgetting and enhances discrimination for both old and new classes.
The remaining rows present the same observations when using the generative replay $\mathcal{G}$.

\vspace{-0.15cm}
\paragraph{Generative replay size.}
Figure~\ref{fig:replay_size} shows the effect of the generative replay size. It shows that 
\mymodel{} consistently outperforms $\mathcal{G}$-SD for different replay size. 
Also, even tiny generated samples from \mymodel{} significantly contribute to increasing the mIoU and hIoU. In contrast, the small size of generated samples~(e.g., 20) from Stable Diffusion even degrades the performance, which also demonstrates the problems of distribution mismatch.

\section{Conclusion and limitations}
We proposed a novel and adaptive generative replay-based framework for the CISS task.
While our framework shows promise, it does have certain limitations. The first limitation is the increased storage requirements due to the need for additional device storage to store the generated old class samples. This can be challenging when dealing with a large number of classes.
Another limitation is the slow inference speed of the text-to-image diffusion model, which hinders real-time or efficient inference when scaling up to a million classes in CISS.
Despite these limitations, our framework still offers valuable contributions in addressing CISS challenges and can be effective in various scenarios.

\bibliographystyle{plain}
\bibliography{reference}







\end{document}